# Study of Residual Networks for Image Recognition

Mohammad Sadegh Ebrahimi
Stanford University
sadegh@stanford.edu

Hossein Karkeh Abadi
Stanford University
hosseink@stanford.edu

## Abstract

*Deep neural networks demonstrate to have a high performance on image classification tasks while being more difficult to train. Due to the complexity and vanishing gradient problem, it normally takes a lot of time and more computational power to train deeper neural networks. Deep residual networks (ResNets) can make the training process faster and attain more accuracy compared to their equivalent neural networks. ResNets achieve this improvement by adding a simple skip connection parallel to the layers of convolutional neural networks. In this project we first design a ResNet model that can perform the image classification task on the Tiny ImageNet dataset with a high accuracy, then we compare the performance of this ResNet model with its equivalent Convolutional Network (ConvNet). Our findings illustrate that ResNets are more prone to overfitting despite their higher accuracy. Several methods to prevent overfitting such as adding dropout layers and stochastic augmentation of the training dataset has been studied in this work.*

## 1. Introduction

In recent years deep convolutional neural networks have achieved a series of breakthroughs in the field of image classifications [6, 10, 7] . Inspired by simple cells and receptive field discoveries in neuroscience by Hubel and Wiesel [5] Deep convolutional neural nets (CNNs) have a layered structure and each layers is consisted of convolutional filters. By convolving these filters with the input image, feature vectors for the next layer are produced and through sharing parameters, they can be learnt quite easily. Early layers in convolutional neural networks represent low level local features such as edges and color contrasts while deeper layers try to capture more complex shapes and are more specific [10]. One can improve the classification performance of CNNs by enriching the diversity and specificity of these convolutional filters through deepening the network [8]. Although deep networks can have better performance in classification most of the times, they are harder to train mainly due to two reasons:

- Vanishing / exploding gradients: sometimes a neuron dies during training process and depending on its activation function it might never come back [2, 3]. This problem can be addressed with initialization techniques that try to start the optimization process with an active set of neurons.

- Harder optimization: when the model introduces more parameters, it becomes more difficult to train the network. This is not simply an overfitting problem, since sometimes adding more layers leads to even more training errors [9]

Therefore deep CNNs, despite of having better classification performance, are harder to train. One effective way to solve these problems suggested in [4] is Residual Networks (ResNets). The main difference in ResNets is that they have shortcut connections parallel to their normal convolutional layers. Contrary to convolution layers, these shortcut connections are always alive and the gradients can easily back propagate through them, which results in a faster training. In this paper we are going to study ResNets and learn more about the correct ways to use them. In section 2 we explain what are the different ways to design a ResNets based on previous works. In section 3 we will describe the tiny ImageNet datast and Torch, the framework we used for our implementations. In section 4 we explain the methods we used to design our networks, the basic block that we employed in all our networks, and the stochastic data augmentation technique we used to prevent overfitting. . In Section 5 we will discuss our results and show how does a ResNet compare to its equivalent ConvNet. In our conclusion in section 6 we will point out a few considerations that must be accounted when designing a ResNet.

## 2. Related Work

There is a simple difference between ResNets and normal ConvNets. The goal is to provide a clear path for gradients to back propagate to early layers of the network. This makes the learning process faster by avoiding vanishing gra-dient problem or dead Neurons. In the main ResNet paper [4] authors have suggested different configurations of ResNets with 18, 34, 50, 101, and 152 layers. One could describe ResNets as multiple basic blocks that are serially connected to each other and there are also shortcut connections parallel to each basic block and it gets added to its output. Figure 1 shows a basic block introduced in [4]. If the input and output size for a basic block are equal the shortcut connection is simply an identity matrix. Otherwise one can use average pooling (for reduction) and zero padding (for enlargement) to adjust the size. [1] has compared different basic blocks for one shortcut connection in ResNets (Figure 1) and shows that adding a parametered layer after addition can undermine ResNet advantages since there is no fast way for gradients to back through propagate anymore. But considering that condition, there is not huge advantage or disadvantage for adding an un-parametered layer like ReLU or dropout after the addition module.

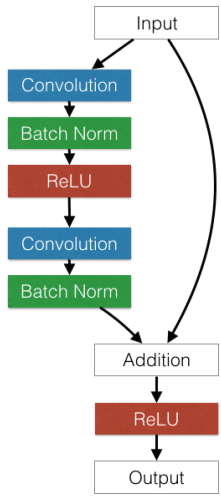

Figure 1. A RestNet basic block

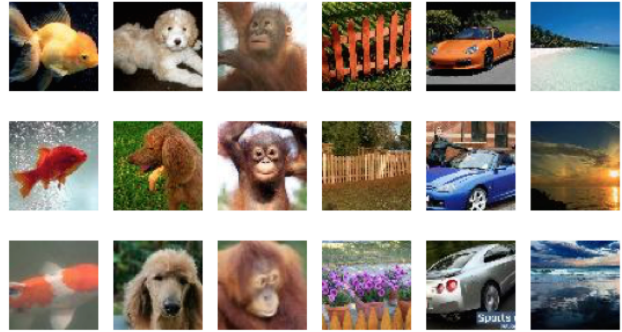

Figure 2. Few sample images from tiny imagenet datasets

## 3. Dataset and implementation

In this section we describe the dataset we worked on and the framework we used for network implementations and model training.

### 3.1. Dataset

In this project we worked on the tiny ImageNet dataset. This dataset consists of a training set of $100,000$ images, a validation set of $10,000$ images, and a test set of $10,000$ images from 200 different classes of objects. All images in tiny ImageNet are $64 \times 64$ and so 4 times smaller than images in the original ImageNet dataset which have a size of $256 \times 256$. Figure 2 shows a few sample images from different classes of tiny imagenet datasets

### 3.2. Torch

Torch is a scientific open source computing framework with wide support for neural network implementaions. In this project we used this framework to implement and train different ResNet and ConvNet Models. Torch has many predefined neural network layers and also packages that enable us to run our training algorithms on GPUs.

## 4. Network Design

The ResNet model introduced in [4] is our starting point for the network design. This model is specifically designed for images in ImageNet and accepts images with size $256 \times 256$ and classifies them in 1000 categories. There are many different methods one can employ to start with this trained model and alter it to accept tiny ImageNet images with size $64 \times 64$ and classify them into 200 categories. A nave method could be just up-sampling a $64 \times 64$ image to a $256 \times 256$ and then give it to the trained model, or just skipping the first layer and insert the original image as the input of the second convolutional layer, and then fine tuning a few of the last layers to get higher accuracy. However, since in this project were interested in comparing ResNet models with their equivalent ConvNets, we had to design and train our models from scratch (although, we might get worse accuracies because of lack of computational resources). In this section we first describe different Networks architectures we designed for image classification task and then we illustrate the stochastic data augmentation we used to prevent the model from overfitting.

### 4.1. Network Architectures

If we train the original 18-layer ResNet introduced in [4] on tiny ImageNet dataset, we wlll see that this model suffers from overfitting. In order to reduce overfitting we introduced a new Basic Block (BB) shown in figure 3 by adding a dropout layer with parameter $0.5$ between the two convolution layers in the basic block shown in figure 1. We used ReLU for the nonlinearity unit in all the neurons.

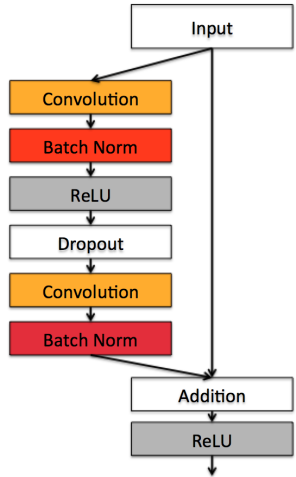

Figure 3. New basic block with a dropout layer to reduce overfitting

Figure 4 shows one of the ResNets we designed for the image classification tasks. This model gives a Top-1 classification accuracy of $49\%$ on validation set of tiny ImageNet. For more details see section 5.

### 4.2. Data Augmentation

There are only 500 images per each class in tiny ImageNet dataset. This makes the tiny ImageNet to be considered as a small dataset if we're training deep neural networks on them. To overcome overfitting, we need an augmentation method to increase the size of the dataset. One method is to add a few cropped version of each image and their horizontal flipped images to the dataset. However, since there is a limited memory space available, now we cannot load all of the images of the new richer dataset into the memory. So, instead of doing all the data augmentations offline, in our implementations we used an online version of it. Whenever a new batch arrives, we pass all images in that batch from a random transformation unit. This unit first flips the image horizontally with probability $0.5$, and then with some probabilty $p$, it randomly crops the image to a $56 \times 56$ image and then rescale it to it's original size, $64 \times 64$. Figure 5 shows how this unit works on sample image. We used $p = 0.7$ in our implementation.

## 5. Experiment Results

As mentioned before, even though vanishing gradient problem is a big issue for deep neural networks, in shallow ConvNets it is not a big deal. In order to observe this effect we compared two shallow networks with 7 and 9 layers. Figure 6 and 7 show the loss function and traing and validation accuracy of these two networks on CIFAR-10 dataset. As we see for 9 layer network ResNet and ConvNet have similar performance and for even shallower networks (7 layers) the ResNet performance is even worse that plain ConvNets. This result makes sense because when you are adding the output of a convolutional layer with its input, you are basically averaging a trained processed data with the raw data and that would just harm the training if there were no other benefit to it. But if there are other benefits to it (for example in deep networks) the overall effects could be improved accuracy.

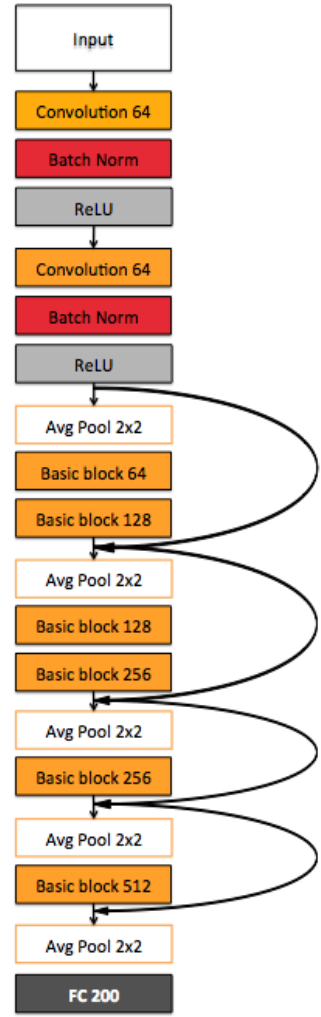

Figure 4. A sample ResNet model for Image Classification

Then we tried to train multiple deep ResNets varying from 12 to 21 layers and see which one performs better on the Tiny Imagenet data set. The results are brought in Table 1. Note that all the models are trained for the same number of epochs. We picked the best performing network (Net 1) with $49\%$ percent validation accuracy and trained an equivalent plain ConvNet with the same architecture (Net 6). In Figure 3 we compared the accuracy of these two networks (ResNet and equivalent ConvNet). One could clearly see that the ResNet has much higher accuracies than plain ConvNet and it trains much faster. In this ResNet the validation accuracy, 13%, and training accuracy 30% percent higher than its ConvNet equivalent. The difference between training accuracy and validation accuracy is a good indicator of over fitting and based on our results we realized that ResNets are more prone to overfitting. In figure 8, one can see that this difference for plain ConvNet is 7% while in ResNet it is around 23%. Figure 9 shows how loss decreased while training both models. Originally this difference was even higher for ResNet (around 30%) but we used dropout and stochastic augmentation technique that was described in section 3 to reduce this overfitting but could only reduce it by 6%. Another way to reduce the overfitting is to have a smaller parameter set which means less convolution layers. In Net 3 (Table 1) we implemented such a network and the training and validation accuracy difference was reduced to 16%, but the down side of this model was to have smaller validation accuracy in compare to best network (by 3%). Both dropout and stochastic augmentation was used in this implementation.

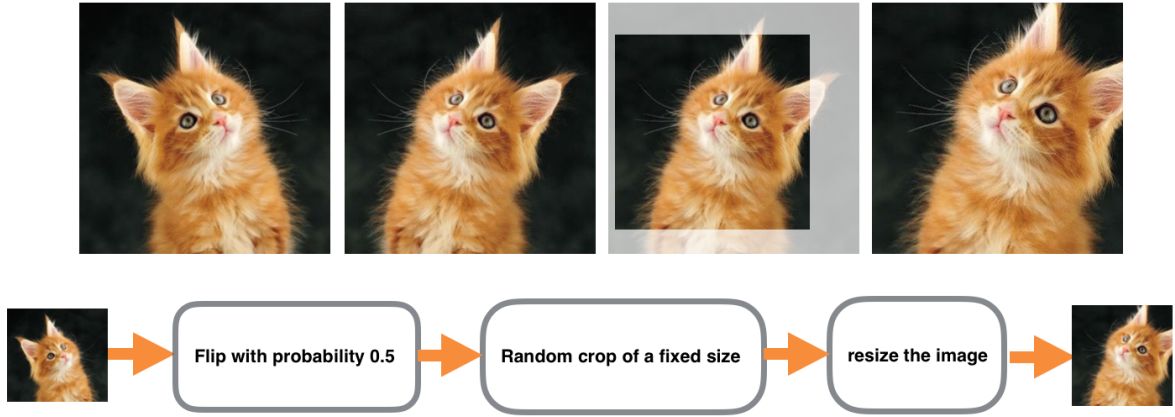

Figure 5. Online data augmentation of a sample image

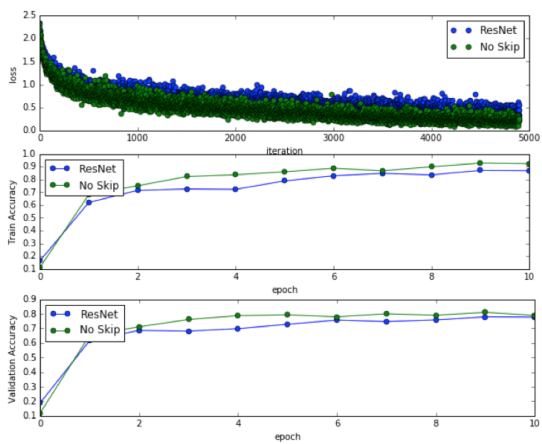

Figure 6. Training/Validation accuracy and loss at each epoch for a 7-layer network over CIFAR-10 dataset

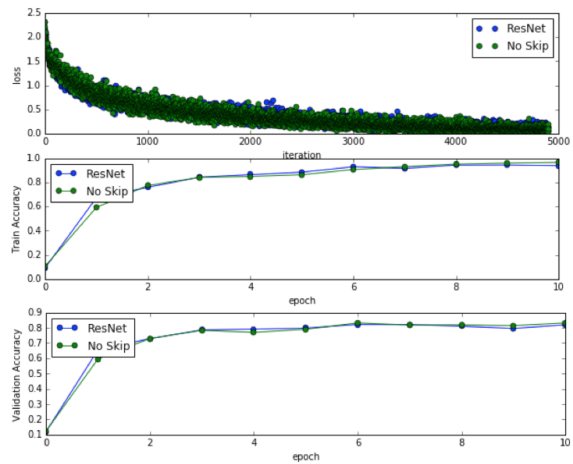

Figure 7. Training/Validation accuracy and loss at each epoch for a 9-layer network over CIFAR-10 dataset

## 6. Conclusion

As we explained in our results adding a simple shortcut connection can improve the accuracy in the image classification task and make the training process much faster. But the trade of is that residual networks are more prone to overfitting which is undesirable. We showed that by using different machine learning techniques like drop out layer and stochastic augmentation we can reduce this overfitting and if designed properly we can have fewer parameters that re-sult in much smaller over fitting (14%). We also observed that resnets are more powerful for very deep networks and if employed improperly it could even hurt the performance for very shallow networks. To conclude Resnets show promising landscape in deep learning but it should not be just blindly used and there is a lot of room to study and understand their functionality and correct use.

|  | Net 1 | Net 2 | Net 3 | Net 4 | Net 5 | Net 6 |
|---|---|---|---|---|---|---|
|  | (Conv 64)× 2 | Conv 64 | Conv 32 | (Conv 64) × 2 | (Conv 64) × 2 | (Conv 64) × 2 |
|  | Avg 2 | Avg 2 | Conv 64 | Avg 2 | Avg 2 | Avg 2 |
|  | BB 64 | BB 128 | Max 2 | (BB 64) ×3 | BB 128 | (Conv 64) × 2 |
|  | BB 128 | Avg 2 | (BB 64) × 2 | Max 2 | Max 2 | (Conv 128) × 2 |
|  | Avg 2 | BB 128 | Avg 2 | (BB 128) × 3 | (BB 256) × 2 | Avg 2 |
|  | BB 128 | BB 256 | (BB 128) × 3 | Max 2 | Max 2 | (Conv 128) × 2 |
|  | BB 256 | Avg 2 | Avg 2 | BB 256 | (BB 512) × 2 | (Conv 256) × 2 |
|  | Avg 2 | BB 256 | BB 128 | BB512 | Avg 2 | Avg 2 |
|  | BB 256 | Avg 2 | Avg 2 | Avg 2 | Dropout | (Conv 256) × 2 |
|  | Avg 2 | BB 512 | BB 256 | Dropout | FC 200 | Avg 2 |
|  | BB 512 | Avg 4 | Avg 2 | FC 200 |  | (Conv 512) × 2 |
|  | Avg 2 | FC 200 | FC 200 |  |  | Avg2 |
|  | FC 200 |  |  |  |  | FC 200 |
| Number of layers | 15 | 12 | 17 | 21 | 15 | 15 |
| Training Accuracy | 72% | 69% | 62% | 44% | 55% | 43% |
| Validation Accuracy | 49% | 46% | 46% | 36% | 42% | 36% |

Table 1. Training and validation accuracies of different models

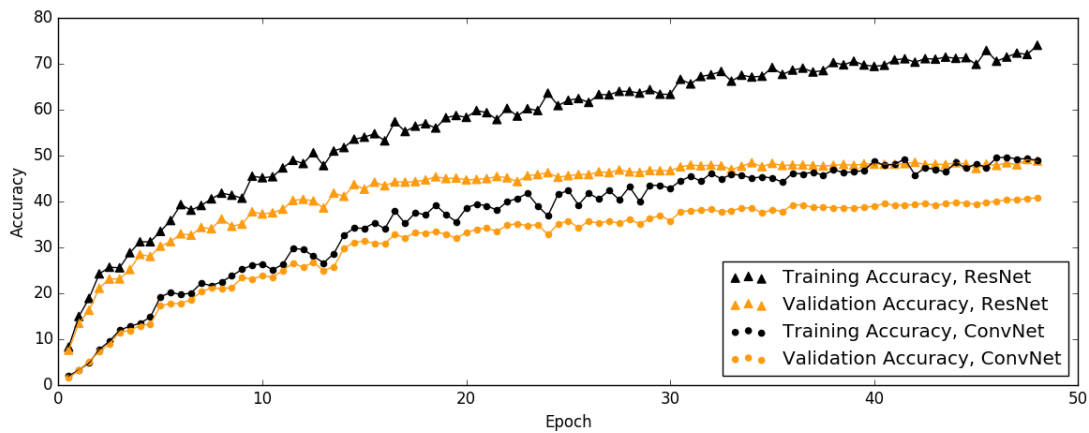

Figure 8. Training and Validation accuracies for ResNet Net1 described in the table 1 and it's equivalent ConvNet.

## References

[1] http://torch.ch/blog/2016/02/04/resnets.html.
[2] Y. Bengio, P. Simard, and P. Frasconi. Learning long-term dependencies with gradient descent is difficult. *Neural Networks, IEEE Transactions on*, 5(2):157–166, 1994.
[3] X. Glorot and Y. Bengio. Understanding the difficulty of training deep feedforward neural networks. In *International conference on artificial intelligence and statistics*, pages 249–256, 2010.
[4] K. He, X. Zhang, S. Ren, and J. Sun. Deep residual learning for image recognition. *arXiv preprint arXiv:1512.03385*, 2015.
[5] D. H. Hubel and T. N. Wiesel. Receptive fields and functional architecture of monkey striate cortex. *The Journal of physiology*, 195(1):215–243, 1968.
[6] A. Krizhevsky, I. Sutskever, and G. E. Hinton. Imagenet classification with deep convolutional neural networks. In *Advances in neural information processing systems*, pages 1097–1105, 2012.
[7] P. Sermanet, D. Eigen, X. Zhang, M. Mathieu, R. Fergus, and Y. LeCun. Overfeat: Integrated recognition, localization and detection using convolutional networks. *arXiv preprint arXiv:1312.6229*, 2013.

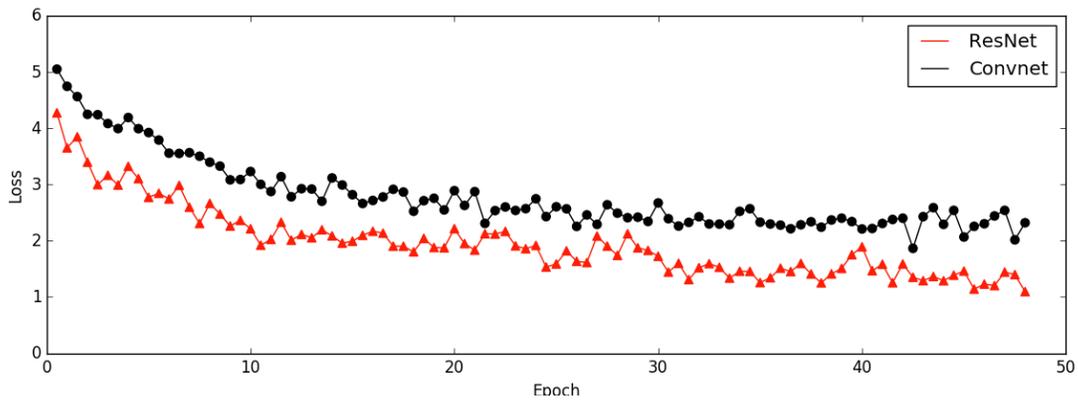

Figure 9. Loss vs. epoch in the training of the ResNet Net1 described in the table 1 and it's equivalent ConvNet.

[8] K. Simonyan and A. Zisserman. Very deep convolutional networks for large-scale image recognition. *arXiv preprint arXiv:1409.1556*, 2014.

[9] R. K. Srivastava, K. Greff, and J. Schmidhuber. Highway networks, 2015.

[10] M. D. Zeiler and R. Fergus. Visualizing and understanding convolutional networks. In *Computer vision–ECCV 2014*, pages 818–833. Springer, 2014.